# Learning Traffic as Images: A Deep Convolutional Neural Network for Large-Scale Transportation Network Speed Prediction

**Xiaolei Ma [1], Zhuang Dai [1], Zhengbing He [2], Jihui Ma [2,\*], Yong Wang [3] and Yunpeng Wang [1]**

[1] School of Transportation Science and Engineering, Beijing Key Laboratory for Cooperative Vehicle Infrastructure System and Safety Control, Beihang University, Beijing 100191, China; xiaolei@buaa.edu.cn (X.M.); zhuangdai@buaa.edu.cn (Z.D.); ypwang@buaa.edu.cn (Y.W.)
[2] School of Traffic and Transportation, Beijing Jiaotong University, Beijing 100044, China; he.zb@hotmail.com
[3] School of Economics and Management, Chongqing Jiaotong University, Chongqing 400074, China; yongwx6@gmail.com
**\*** Correspondence: jhma@bjtu.edu.cn; Tel.: +86-10-5168-8514

**Abstract:** This paper proposes a convolutional neural network (CNN)-based method that learns traffic as images and predicts large-scale, network-wide traffic speed with a high accuracy. Spatiotemporal traffic dynamics are converted to images describing the time and space relations of traffic flow via a two-dimensional time-space matrix. A CNN is applied to the image following two consecutive steps: abstract traffic feature extraction and network-wide traffic speed prediction. The effectiveness of the proposed method is evaluated by taking two real-world transportation networks, the second ring road and north-east transportation network in Beijing, as examples, and comparing the method with four prevailing algorithms, namely, ordinary least squares, k-nearest neighbors, artificial neural network, and random forest, and three deep learning architectures, namely, stacked autoencoder, recurrent neural network, and long-short-term memory network. The results show that the proposed method outperforms other algorithms by an average accuracy improvement of 42.91% within an acceptable execution time. The CNN can train the model in a reasonable time and, thus, is suitable for large-scale transportation networks.

**Keywords:** transportation network; traffic speed prediction; spatiotemporal feature; deep learning; convolutional neural network

## 1. Introduction

Predicting the future is one of the most attractive topics for human beings, and the same is true for transportation management. Understanding traffic evolution for the entire road network rather than on a single road is of great interest and importance to help people with complete traffic information in make better route choices and to support traffic managers in managing a road network and allocate resources systematically [1,2].

However, large-scale network traffic prediction requires more challenging abilities for prediction models, such as the ability to deal with higher computational complexity incurred by the network topology, the ability to form a more intelligent and efficient prediction to solve the spatial correlation of traffic in roads expanding on a two-dimensional plane, and the ability to forecast longer-term futures to reflect congestion propagation. Unfortunately, traditional traffic prediction models, which usually treat traffic speeds as sequential data, do not provide those abilities because of limitations, such as hypotheses and assumptions, ineptness to deal with outliers, noisy or missing data, and inability to cope with the curse of dimensionality [3]. Thus, existing models may fail to predict large-scale network traffic evolution.

In the existing literature, two families of research methods have dominated studies in traffic forecasting: statistical methods and neural networks [3].

Statistical techniques are widely used in traffic prediction. For example, according to the periodicity of traffic evolutions, nonparametric models, such as k-nearest neighbors (KNN), have been applied to predict traffic speeds and volumes [4–6]. More advanced models were employed, including support vector machines (SVM) [7], seasonal SVM [8], Online-SVM [9], and on-line sequential extreme learning machine [10], to promote prediction accuracy by capturing the high dynamics and sensitivity of traffic flow. SVM performance in large-scale traffic speed prediction was further improved [8,11]. Multivariate nonparametric regression was also used in traffic prediction [12,13]. Recently, a wealth of literature leverage multiple hybrid models and spatiotemporal features to improve traffic prediction performance. For example, Li et al. [14] proposed a hybrid strategy with ARIMA and SVR models to enhance traffic prediction power by considering both spatial and temporal features. Zhu et al. [15] employed a linear conditional Gaussian Bayesian network (LCG-BN) with spatial and temporal, as well as speed, information for traffic flow prediction. Li et al. [16] studied the chaotic situation of traffic flow based on a Bayesian theory-based prediction algorithm, and incorporated speed, occupancy, and flow for accuracy improvement. Considering the correlations shown in successive time sequences of traffic variables, time-series prediction models have been widely employed in traffic prediction. One of the typical models is the autoregressive integrated moving average (ARIMA) model, which considers the essential traffic flow characteristics, such as inherent correlations (via a moving average) and its effect on the short future (via autoregression). To date, the model, and its extensions, such as the seasonal ARIMA model [17,18], KARIMA model [19], and the ARIMAX model [20], have been widely studied and applied. In summary, statistical methods have been widely used in traffic prediction, and promising results have been demonstrated. However, these models ignore the important spatiotemporal feature of transportation networks, and cannot be applied to predict overall traffic in a large-scale network. SVM usually takes a long time and consumes considerable computer memory on training and, hence, it might be powerless in large data-related applications.

Artificial neural networks (ANNs) are also usually applied to traffic prediction problems because of its advantages, such as their capability to work with multi-dimensional data, implementation flexibility, generalizability, and strong forecasting power [3]. For example, Huang and Ran [21] used an ANN to predict traffic speed under adverse weather conditions. Park et al. [2] presented a real-time vehicle speed prediction algorithm based on ANN. Zheng et al. [22] combined an ANN with Bayes' theorem to predict short-term freeway traffic flow. Moretti et al. [23] developed a statistical and ANN bagging ensemble hybrid model to forecast urban traffic flow.

However, the data-driven mechanism of an ANN cannot explain the spatial correlations of a road network particularly well. In addition, compared with deep learning approaches, the prediction accuracy of an ANN is lower because of its shallow architecture. Recently, more advanced and powerful deep learning models have been applied to traffic prediction. For example, Polson and Sokolov [24] used deep learning architectures to predict traffic flow. Huang et al. [25] first introduced Deep Belief Networks (DBN) into transportation research. Then, Tan et al. [26] compared the performance of DBNs with two kinds of RBM structures, namely, RBM with binary visible and hidden units (B-B RBM) and RBM with Gaussian visible units and binary hidden units (G-B RBM), and found that the former outperforms the later in traffic flow prediction. Ma et al. [27] combined deep restricted Boltzmann machines (RBM) with a recurrent neural network (RNN) and formed a RBM-RNN model that inherits the advantages of both RBM and RNN. Lv et al. [28] proposed a novel deep-learning-based traffic prediction model that considered spatiotemporal relations, and employed stack autoencoder (SAE) to extract traffic features. Duan et al. [29] used denoising stacked autoencoders (DSAE) for traffic data imputation. Ma et al. [30] introduced a long short-term memory neural network (LSTM NN) into traffic prediction and demonstrated that LSTM NN outperformed other neural networks in both stability and accuracy in terms of traffic speed prediction by using remote microwave sensor data collected from the Beijing road network.

Deep learning methods exploit much deeper and more complex architectures than an ANN, and can achieve better results than traditional methods. However, these attempts still mainly focus on the prediction of traffic on a road section or a small network region. Few studies have considered a transportation network as a whole and directly estimated the traffic evolution on a large scale. More importantly, the majority of these models merely considered the temporal correlations of traffic evolutions at a single location, and did not consider its spatial correlations from the perspective of the network.

To fill the gap, this paper introduces an image-based method that represents network traffic as images, and employs the deep learning architecture of a convolutional neural network (CNN) to extract spatiotemporal traffic features contained by the images. A CNN is an efficient and effective image processing algorithm and has been widely applied in the field of computer vision and image recognition with remarkable results achieved [31,32]. Compared with prevailing artificial neural networks, a CNN has the following properties in extracting features: First, the convolutional layers of a CNN are connected locally instead of being fully connected, meaning that output neurons are only connected to its local nearby input neurons. Second, a CNN introduces a new layer-construction mechanism called pooling layers that merely select salient features from its receptive region and tremendously reduce the number of model parameters. Third, normal fully-connected layers are used only in the final stage, when the dimension of input layers is controllable. The locally-connected convolutional layers enable a CNN to efficiently deal with spatially-correlated problems [31,33,34]. The pooling layers makes CNNs generalizable to large-scale problems [35]. The contributions of the paper can be summarized as follows:

- The temporal evolutions and spatial dependencies of network traffic are considered and applied simultaneously in traffic prediction problems by exploiting the proposed image-based method and deep learning architecture of CNNs.
- Spatiotemporal features of network traffic can be extracted using a CNN in an automatic manner with a high prediction accuracy.
- The proposed method can be generalized to large-scale traffic speed prediction problems while retaining trainability because of the implementation of convolutional and pooling layers.

The rest of the paper is organized as follows: In Section 2, a two-step procedure that includes converting network traffic to images and a CNN for network traffic prediction is introduced. In Section 3, four prediction tests are conducted on two transportation networks using the proposed method, and are compared with the other prevailing prediction methods. Finally, conclusions are drawn with future study directions in Section 4.

## 2. Methods

Traffic information with time and space dimensions should be jointly considered to predict network-wide traffic congestion. Let *x*- and *y*-axis represent time and space of a matrix, respectively. The elements within the matrix are values of traffic variables associated with time and space. The generated matrix can be viewed as a channel of an image in the way that every pixel in the image shares the corresponding value in the matrix. As a result, the image is of *M* pixels width and *N* pixels height, where *M* and *N* are the two dimensions of the matrix. A two-step methodology, converting network traffic to images and the CNN for network traffic prediction, respectively, is designed to learn from the matrix and make predictions.

*2.1. Converting Network Traffic to Images*

A vehicle trajectory recorded by a floating car with a dedicated GPS device provides specific information on vehicle speed and position at a certain time. From the trajectory, the spatiotemporal traffic information on each road segment can be estimated and integrated further into a time-space matrix that serves as a time-space image.

In the time dimension, time usually ranges from the beginning to the end of a day, and time intervals, which are usually 10 s to 5 min, depend on the sampling resolution of the GPS devices. Generally, narrow intervals, for example 10 s, are meaningless for traffic prediction. Thus, if the sampling resolution is high, these data may be aggregated to obtain wider intervals, such as several minutes.

In the space dimension, the selected trajectory is viewed as a sequence of dots with inner states, including vehicle position, average speed, etc. This sequence of dots can be ordered simply and linearly fitted into the *y*-axis, but may result in a high dimension and uninformative issues, because the sequences of dots are redundant and a large number of regions in this sequence are stable and lack variety. Therefore, to make the *y*-axis both compact and informative, the dots are grouped into sections, each representing a similar traffic state. The sections are then ordered spatially with reference to a predefined start point of a road, and then fitted into the *y*-axis.

Finally, a time-space matrix can be constructed using time and space dimension information. Mathematically, we denote the time-space matrix by:

$$M = \begin{bmatrix} m_{11}, m_{12}, \cdots, m_{1N} \\ m_{21}, m_{22}, \cdots, m_{2N} \\ \vdots & \vdots & \cdots & \vdots \\ m_{Q1}, m_{Q2}, \cdots, m_{QN} \end{bmatrix} \quad (1)$$

where *N* is the length of time intervals, *Q* is the length of road sections; the *i*th column vector of *M* is the traffic speed of the transportation network at time *i*; and pixel $m_{ij}$ is the average traffic speed on section *i* at time *j*. Matrix *M* forms a channel of the image. Figure 1 illustrates the relations among raw averaged floating car speeds, time-space matrix, and the final image.

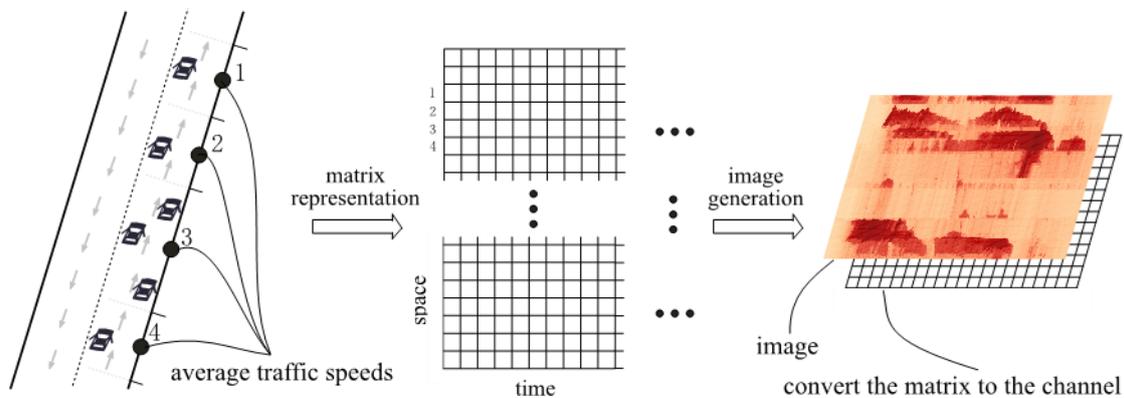

**Figure 1.** An illustration of the traffic-to-image conversion on a network.

*2.2. CNN for Network Traffic Prediction*

2.2.1. CNN Characteristics

The CNN has exhibited a significant learning ability in image understanding because of its unique method of extracting critical features from images. Compared to other deep learning architectures, two salient characteristics contribute to the uniqueness of CNN, namely, (a) locally-connected layers, which means output neurons in the layers are connected only to their local nearby input neurons, rather than the entire input neurons in fully-connected layers. These layers can extract features from an image effectively, because every layer attempts to retrieve a different feature regarding the prediction problem [31]; and (b) a pooling mechanism, which largely reduces the number of parameters required to train the CNN while guaranteeing that the most important features are preserved.

Sharing the two salient characteristics, the CNN is modified in the following aspects to adapt to the context of transportation: First, the model inputs are different, i.e., the input images have only one channel valued by traffic speeds of all roads in a transportation network, and the pixel values in the images range from zero to the maximum traffic speed or speed limits of the network. In contrast, in the image classification problem, the input images commonly have three channels, i.e., RGB, and pixel values range from 0 to 255. Although differences exist, the model inputs are normalized to prevent model weights from increasing the model training difficulty. Second, the model outputs are different. In the context of transportation, the model outputs are predicted traffic speeds on all road sections of a transportation network, whereas, in the image classification problem, model outputs are image class labels. Third, abstract features have different meanings. In the context of transportation, abstract features extracted by the convolutional and pooling layers are relations among road sections regarding traffic speeds. In the image classification problem, the abstract features can be shallow image edges and deep shapes of some objects in terms of its training objective. All of these abstract features are significant for a prediction problem [36]. Fourth, the training objectives differ because of distinct model outputs. In the context of transportation, because the outputs are continuous traffic speeds, continuous cost functions should be adopted accordingly. In the image classification problem, cross-entropy cost functions are usually used.

2.2.2. CNN Characteristics

Figure 2 shows the structure of CNN in the context of transportation with four main parts, that is, model input, traffic feature extraction, prediction, and model output. Each of the parts is explained below.

First, model input is the image generated from a transportation network with spatiotemporal characteristics. Let the lengths of input and output time intervals be $F$ and $P$, respectively. The model input can be written as:

$$x^i = [m_i, m_{i+1}, ..., m_{i+P-1}], \ i \in [1, N-P-F+1] \quad (2)$$

where $i$ is the sample index, $N$ is the length of time intervals, and $m_i$ is a column vector representing traffic speeds of all roads in a transportation network within one time unit.

Second, the extraction of traffic features is the combination of convolutional and pooling layers, and is the core part of the CNN model. The pooling procedure is indicated by using *pool*, and $L$ is denoted by the depth of CNN. Denote the input, output, and parameters of $l$th layer by $x_l^j$, $o_l^j$ and $(W_l^j, b_l^j)$, respectively, where $j$ is the channel index considering the multiple convolutional filters in the convolutional layer. The number of convolutional filters in $l$th layer is denoted by $c_l$. The output in the first convolutional and pooling layers can be written as:

$$o_1^j = pool\left(\sigma\left(W_1^j x_1^j + b_1^j\right)\right), \ j \in [1, c_1] \quad (3)$$

where $\sigma$ is the activation function, which will be discussed in next section. The output in the $l$th ($l \neq 1$, $l = 1\ L$) convolutional and pooling layers can be written as:

$$o_l^j = pool\left(\sigma\left(\sum_{k=1}^{c_{l-1}}\left(W_l^j x_l^k + b_l^j\right)\right)\right), \ j \in [1, c_l] \quad (4)$$

The extraction of traffic features has the following characteristics: (a) Convolution and pooling are processed in two dimensions. This part can learn the spatiotemporal relations of the road sections in terms of the prediction task in model training; (b) Different from layers with only four convolutions or pooling filters in Figure 2, in reality, the number of the layers in applications are set to be hundreds, which means hundreds of features can be learned by a CNN; and (c) a CNN transforms the model input into deep features through these layers.

In the model prediction, the features learnt and outputted by traffic feature extraction are concatenated into a dense vector that contains the final and most high-level features of the input transportation network. The dense vector can be written as:

$$o_L^{flatten} = flatten\left(\left[o_L^1, o_L^2, ..., o_L^j\right]\right), j = c_L \tag{5}$$

where *L* is the depth of CNN and *flatten* is the concatenating procedure discussed above.

Finally, the vector is transformed into model outputs through a fully connected layer. The model output can, thus, be written as:

$$\begin{aligned}\hat{y} &= W_f o_L^{flatten} + b_f \\ &= W_f \left( flatten\left( pool\left( \sigma\left( \sum_{k=1}^{c_{l-1}} \left(W_L^j x_L^k + b_L^j\right)\right)\right)\right)\right) + b_f\end{aligned} \tag{6}$$

where $W_f$ and $b_f$ are parameters of the fully connected layer. $\hat{y}$ are the predicted network-wide traffic speeds.

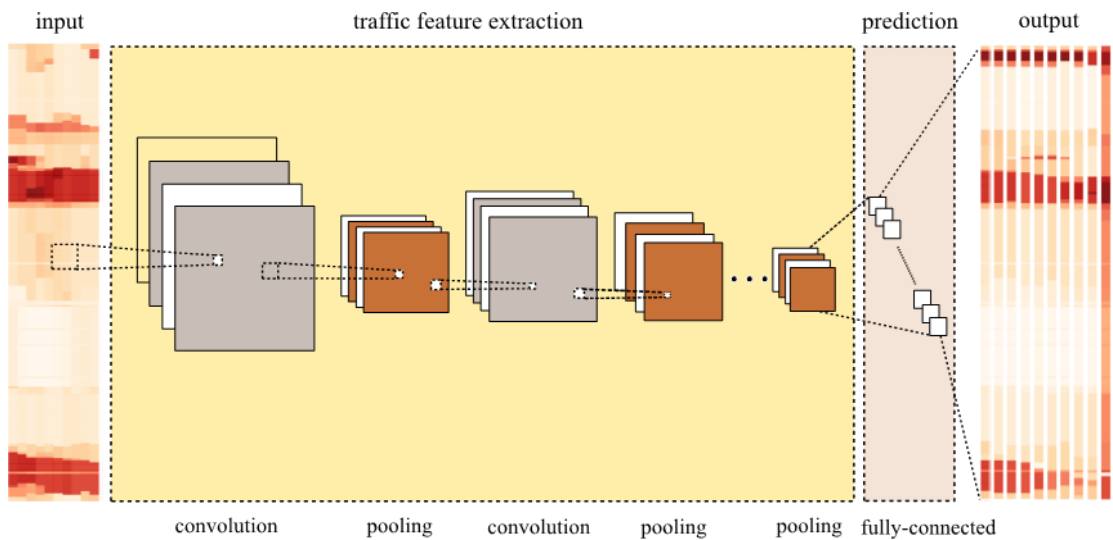

**Figure 2.** Deep learning architecture of CNN in the context of transportation.

2.2.3. Convolutional Layers and Pooling Layers of the CNN

Before discussing the explicit layers, it should be noted that each layer is activated by an activation function. The benefits of employing the activation function are as follows: (a) the activation function transforms the output to a manageable and scaled data range, which is beneficial to model training; and (b) the combination of the activation function through layers can mimic very complex nonlinear functions making the CNN powerful enough to handle the complexity of a transportation network. In this study, the Relu function is applied and defined as follows:

$$g_1(x) = \begin{cases} x, & \text{if } x > 0 \\ 0, & \text{otherwise} \end{cases} \tag{7}$$

Convolutional layers differ from traditional feedforward neural network where each input neuron is connected to each output neuron and the network is fully connected (fully-connected layer). The CNN uses convolutional filters over its input layer and obtains local connections where only local input neurons are connected to the output neuron (convolutional layer). Hundreds of filters are sometimes applied to the input and results are merged in each layer. One filter can extract one traffic feature from the input layer and, thus, hundreds of filters can extract hundreds of traffic features. Those extracted traffic features are combined further to extract a higher level and more abstract traffic features. The process confirms the compositionality of the CNN, meaning each filter composes a local path from lower-level into higher-level features. When one convolutional filter $W_l^r$ is applied to the input, the output can be formulated as:

$$y_{conv} = \sum_{e=1}^{m}\sum_{f=1}^{n}\left(\left(W_l^r\right)_{ef} d_{ef}\right) \tag{8}$$

where $m$ and $n$ are two dimensions of the filter, $d_{ef}$ is the data value of the input matrix at positions $e$ and $f$, and $\left(W_l^r\right)_{ef}$ is the coefficient of the convolutional filter at positions $e$ and $f$ and $y_{conv}$ is the output.

Pooling layers are designed to downsample and aggregate data because they only extract salient numbers from the specific region. The pooling layers guarantee that CNN is locally invariant, which means that the CNN can always extract the same feature from the input, regardless of feature shifts, rotations, or scales [36]. Based on the above facts, the pooling layers can not only reduce the network scale of the CNN, but also identify the most prominent features of input layers. Taking the maximum operation as an example, the pooling layer can be formulated as:

$$y_{pool} = \max\left(d_{ef}\right), \ e \in [1 \cdots p], \ f \in [1 \cdots q] \tag{9}$$

where $p$ and $q$ are two dimensions of pooling window size, $d_{ef}$ is the data value of the input matrix at positions $e$ and $f$, and $y_{pool}$ is the pooling output.

2.2.4. CNN Optimization

The predictions of the CNN are traffic speeds on different road sections, and the mean squared errors (MSEs) are employed to measure the distance between predictions and ground-truth traffic speeds. Thus, minimizing MSEs is taken as the training goal of the CNN. MSE can be written as:

$$MSE = \frac{1}{n}\sum_{i=1}^{N}\left(\hat{y}_i - y_i\right)^2 \tag{10}$$

Let the model parameters be set $\Theta = \left(W_l^i, b_l^i, W_f, b_f\right)$, the optimal values of $\Theta$ can be determined according to the standard backpropagation algorithm similar to other studies on CNN [31,36]:

$$\begin{aligned}\Theta &= \arg\min_{\Theta}\frac{1}{n}\sum_{i=1}^{N}\left(\hat{y}_i - y_i\right)^2 \\ &= \arg\min_{\Theta}\frac{1}{n}\left\|W_f o_L^{flatten} + b_f - y\right\|^2 \\ &= \arg\min_{\Theta}\frac{1}{n}\left\|W_f\left(flatten\left(pool\left(\sigma\left(\sum_{k=1}^{c_{l-1}}\left(W_L^j x_L^k + b_L^j\right)\right)\right)\right)\right) + b_f - y\right\|^2\end{aligned} \tag{11}$$

3. Empirical Study

*3.1. Data Description*

Beijing is the capital of China and one of the largest cities in the world. At present, Beijing is encircled by four two-way ring roads, that is, the second to fifth ring roads, and has about 10,000 taxis to serve its population of more than 21 million. These taxis are equipped with GPS devices that upload data approximately every minute. The uploaded data contain information, including car positions, recording time, moving directions, vehicle travel speeds, etc. The data were collected from 1 May 2015 to 6 June 2015 (37 days). These data are well-qualified probe data because the missing data accounts for less than 2.9%, and are properly remedied using spatiotemporal adjacent records. In this paper, data are aggregated into two-min intervals because data usually fluctuated over shorter time intervals, and the aggregation will cause data to be more stable and representative.

In this paper, two sub-transportation networks, i.e., the second ring (labeled as Network 1) and north-east transportation network (labeled as Network 2) of Beijing, are selected to demonstrate the proposed method. The two networks differ in network size and topology complexity, as shown in

Figure 3. Network 1 consists of 236 road sections for aggregating GPS data, all of which are one-way roads. Network 2 consists of 352 road sections, including two-way and crossroads. The selected networks represent different road topologies and structures and, thus, can be used to better evaluate the effectiveness of the proposed CNN traffic prediction algorithm.

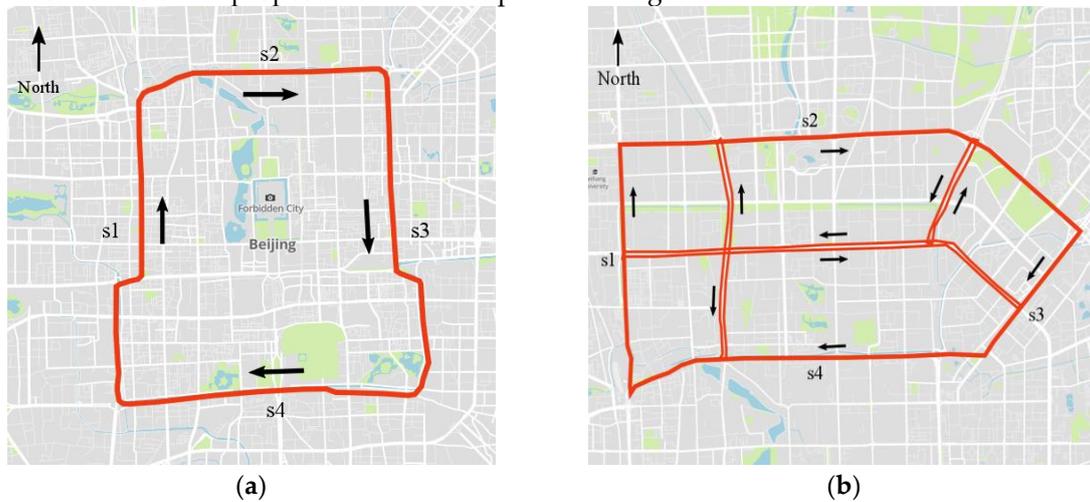

**Figure 3.** Two sub-transportation networks for testing: (**a**) Network 1, the second ring of Beijing; and (**b**) Network 2, a network in Northeast Beijing

Four prediction tasks are performed to test the CNN algorithm in predicting network-wide traffic speeds. These tasks differ in prediction time spans, i.e., short-term and long-term predictions, and in input information, i.e., prediction using abundant information and prediction using limited information. The four tasks are listed as follows:

Task 1: 10-min traffic prediction using last 30-min traffic speeds;
Task 2: 10-min traffic prediction using last 40-min traffic speeds;
Task 3: 20-min traffic prediction using last 30-min traffic speeds; and
Task 4: 20-min traffic prediction using last 40-min traffic speeds.

In the four tasks, the capabilities and effectiveness of CNN in predicting large-scale transportation network speed can be validated by calculating and comparing the MSEs of CNN.

*3.2. Time-Space Image Generation*

In terms of time-space matrix representation, the goal is to transform spatial relations of the traffic in a transportation network into linear representations. The matrix is straightforward in Network 1 because connected road sections in the ring road can be easily straightened. For Network 2, straightening the road sections into a straight line while maintaining the complete spatial relations of these sections is impossible. A compromise is to segment the network into straight lines and lay road sections in order on these lines. Consequently, in Network 2, only a linear spatial relation on straight lines can be captured. However, complex and network-wide relations of traffic speeds in Network 2 can still be learned because the CNN can learn features from local connections and compose these features into high-level representations [32,36]. Regarding Network 2, the CNN learns the relations of traffic roads from segmented road sections and composes these relations into network-wide relations.

After using a time-space matrix as the channel of an image and representing everyday traffic speeds of the network in an image, 37 images, each corresponding to a day, can be generated for Networks 1 and 2, respectively. Sample images of Networks 1 and 2 on 26 May 2015 are shown in Figure 4. The y-labels of Figure 4, i.e., s1, s2, s3, s4, and other, are road sections shown in Figure 3. The images show rich traffic information, such as the most congested traffic areas, in red regions, and typical congestion propagation patterns, i.e., oscillating congested traffic (OCT) and pinned localized clusters (PLC). A more specific explanation on these traffic patterns can be found in the study by Schönhof and Helbing [37]. Such rich information cannot be well learned by a simple ANN. Thus, a more effective algorithm is necessary.

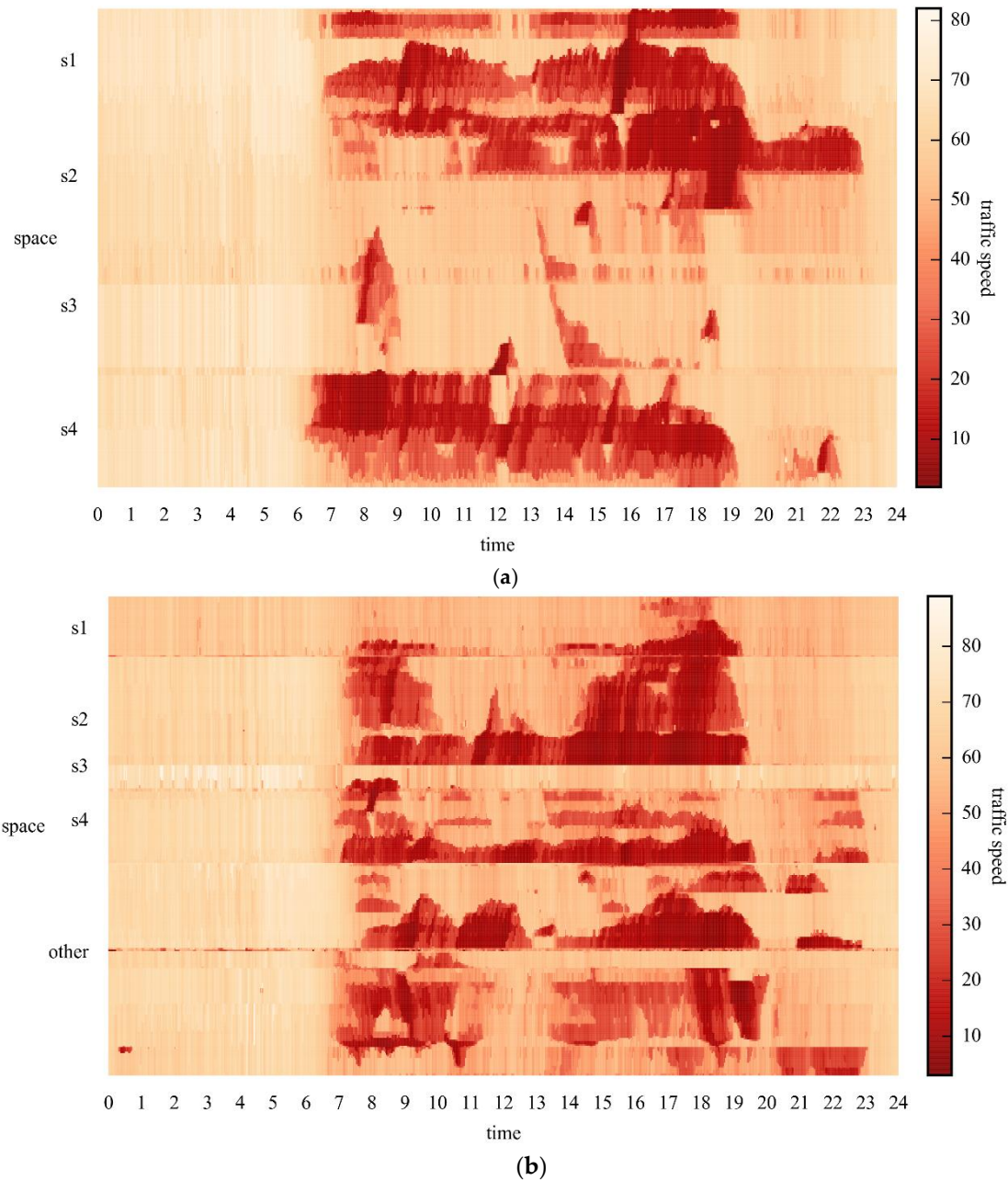

**Figure 4.** Sample images with spatiotemporal traffic speeds for (**a**) Network 1; and (**b**) Network 2.

*3.3. Tuning Up CNN Parameters*

Two critical factors should be considered when implementing the structure of a CNN: (a) hyperparameters concerned with convolutional and pooling layers, such as convolutional filter size, polling size, and polling method; and (b) depth of the CNN.

First, the selection of hyperparameters relies on experts' experience. No general rules can be applied directly. Two well-known examples can be referred. One is LeNet, which marked the beginning of the development of CNN [38], and the other is AlexNet, which won the image classification competition ImageNet in 2010 [31]. Based on the parameter settings of LeNet and AlexNet, we select convolutional filters of size (3, 3) and max poolings of size (2, 2) for the example networks.

Second, the depth of CNN should be neither too large nor too small [39] and, thus, CNN is capable of learning much more complex relations while maintaining the convergence of the model. Different values, from small to large, are assigned to test the CNN model until the incremental benefits are diminished and the convergence becomes difficult in determining a proper value for the depth of the model. The structures of the CNN in different depths are listed in Table 1, where each convolutional layer is followed by a pooling layer, and the numbers represent quantities of convolutional filters in the layer. Obviously, the depth-1 network is a fully connected layer that transforms inputs into predictions, whereas the three other networks first extract spatiotemporal traffic features from the input image using convolutional and pooling layers, and then make predictions based on them. In the experiments, the 40 min historical traffic speeds are used to predict the following 10 min traffic speeds. In model training, 21,600 samples on the first 30 days are used, and in model validation, 5040 samples in the following seven days are used. The results are shown as Figure 5, which shows that adding depth to the CNN model significantly reduces MSEs on the testing data. As a result, a depth-4 CNN model achieves the lowest MSEs on the training and testing data, which are 21.3 and 35.5, respectively. Therefore, the depth-4 model is adopted for experiments in this paper.

**Table 1.** Different depths for CNN.

| Depth | Structures of Prediction Model |
|---|---|
| Depth-1 | A fully connected layer simply makes predictions using the input layer |
| Depth-2 | 64 conv → fully connected |
| Depth-3 | 128 conv → 64 conv → fully connected |
| Depth-4 | 256 conv → 128 conv → 64 conv → fully connected |

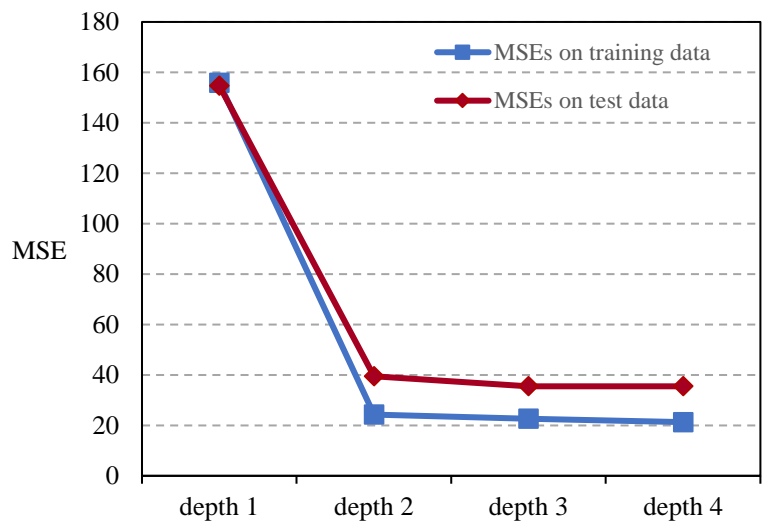

**Figure 5.** Results of CNN in different depths.

The details of the depth-4 CNN are listed in Table 2. The model input has three dimensions (1, 236, 20), where the first number indicates that the input image has one channel, the second number represents the total number of road sections in Network 1, and the third number refers to the input time span, which is 20 time units. Convolutional layers consecutively transform the number of channels into 256, 128, and 64 with the corresponding quantity of convolutional filters, respectively. At the same time, pooling layers consecutively downsample the input window to (118, 10), (59, 5), and (30, 3). The output dimensions in layer 6 are (64, 30, 3), which are then flattened into a vector with a dimension of 5760. The vector is finally transformed into the model output with a dimension of 1180 through a fully-connected layer.

Table 2. Hyperparameters of the CNN.

| Layer | Name | Parameters | Dimensions | Parameter Scale |
|---|---|---|---|---|
| Input | — | — | (1, 236, 20) | — |
| Layer 1 | Convolution | Filter (256, 3, 3) | (256, 236, 20) | 2304 |
| Layer 1 | Pooling | Pooling (2, 2) | (256, 118, 10) | 0 |
| Layer 2 | Convolution | Filter (128, 3, 3) | (128, 118, 10) | 1152 |
| Layer 2 | Pooling | Pooling (2, 2) | (128, 59, 5) | 0 |
| Layer 3 | Convolution | Filter (64, 3, 3) | (64, 59, 5) | 576 |
| Layer 3 | Pooling | Pooling (2, 2) | (64, 30, 3) | 0 |
| Layer 4 | Data flatten | — | (5760, ) | 0 |
| Layer 4 | Fully-connected | — | (1180, ) | 6,796,800 |
| Output | — | — | (1180, ) | — |

Early stopping criterion is applied to prevent the model from overfitting. Model overfitting is a situation where model training does not improve prediction accuracy of the CNN on validation data, although it improves the prediction accuracy of the CNN on testing data. The model should stop training when it begins to overfit. Early stopping is the most common and effective procedure to avoid overfitting issues [40]. This method works in the phase of model training, and early stopping occurrence records losses of the model on the validation dataset. After model training in each epoch, it checks if the losses increase or remain unchanged. Finally, if true and no sign of improvements are observed within a specific number of epochs, model training will be terminated.

*3.4. Results and Comparison*

In order to test the performance of the proposed algorithm, four prevailing statistical algorithms and three deep learning based algorithms are chosen for comparison. OLS is the basic regression algorithm and taken as the benchmark. KNN performs regression using the nearest points. Random forest (RF) makes predictions based on branches of decision trees. ANN represents the traditional neural network and attempts to learn features through hidden layers. SAE is a neural network consisting of multiple layers of autoencoders, where model inputs are encoded into dense or sparse representations before being fed into the next layer [28]. RNN can learn the features by unfolding the time series and capturing the pattern through its shared parameters and hidden states at each time step [27]. LSTM NN is an extension of RNN and becomes popular since the architecture can deal with long-term memories and avoid vanishing gradient issues that traditional RNNs suffer from [30]. These algorithms differ in their ability to predict traffic speeds for multiple road sections in a network. OLS, KNN, and RF can only output the traffic prediction on each link at a time. Hence, to predict network-wide traffic speeds, a large number of models have to be developed. In contrast, ANN, SAE, RNN and LSTM NN can yield network-wide traffic speeds in one model with multi-step outputs. As for the ability to take spatial relations into account, all algorithms treat traffic speeds in different sections as independent sequences and cannot learn spatial relations among sections. Moreover, KNN is configured to use the 10 nearest points. RF is set up to generate 10 decision trees. ANN, RNN, and LSTM NN are optimized to contain three hidden layers with 1000 hidden units in each layer. SAE is tuned up to form up three autoencoder layers with 3000, 2500, and 2000 hidden units in the three layers, respectively.

Table 3 and Figure 6 show the results of different algorithms and CNN when applied to Networks 1 and 2 in four different prediction tasks. The results show that, in all circumstances, the CNN algorithm outperforms other algorithms on testing data, implying that CNN can be better generalized to new data samples. One possible reason is that OLS, KNN, RF, and ANN treat traffic speeds in each section as independent sequences and assumes that traffic speeds in each section are self-affected. This assumption ignores spatial relations among road sections in the network and neglects the important mutual effect of adjacent sections or deeper traffic features. The existing deep learning architectures, i.e. SAE, RNN, and LSTM NN, are also inferior to CNN. This is probably because the majority of existing deep learning-based traffic prediction algorithms cannot incorporate spatial information from the perspective of a network, whereas there exists a strong correlation between multiple congestion bottlenecks [41].

Long-term predictions using CNN can also be validated by comparing the results of tasks 1–4. Usually, when the input time-span is fixed, long-term predictions achieve higher MSEs than short-term predictions, which implies that making long-term predictions is more difficult than making short-term predictions.

**Table 3.** Prediction performance (MSE) of the CNN and other algorithms.

| Study Network | Model | MSE of Different Models (on Test Datasets) | | | |
|---|---|---|---|---|---|
| | | Task 1 | Task 2 | Task 3 | Task 4 |
| Network 1 | CNN | **22.825 *** | **24.345 *** | **30.593 *** | **31.424 *** |
| | OLS | 27.047 | 31.273 | 41.334 | 48.107 |
| | KNN | 51.700 | 55.708 | 60.256 | 64.132 |
| | RF | 35.092 | 35.431 | 40.476 | 40.638 |
| | ANN | 67.764 | 52.339 | 58.797 | 57.225 |
| | SAE | 60.751 | 69.082 | 65.292 | 68.326 |
| | RNN | 33.408 | 36.833 | 40.551 | 39.038 |
| | LSTM NN | 37.759 | 33.218 | 42.909 | 42.865 |
| Network 2 | CNN | **27.163 *** | **28.479 *** | **37.987 *** | **38.816 *** |
| | OLS | 33.741 | 41.657 | 50.123 | 62.282 |
| | KNN | 69.965 | 74.863 | 79.367 | 83.881 |
| | RF | 48.603 | 48.946 | 52.676 | 53.067 |
| | ANN | 124.937 | 147.489 | 133.299 | 168.136 |
| | SAE | 85.079 | 94.982 | 82.271 | 99.020 |
| | RNN | 48.877 | 47.470 | 52.577 | 52.114 |
| | LSTM NN | 43.304 | 45.657 | 50.928 | 48.345 |

Note: * indicates the best result.

We further converted the predicted traffic speeds into three categories of traffic states: heavy traffic (0–20 km/h), moderate traffic (20–40 km/h), and free-flow traffic (>40 km/h). Such a presentation is preferable for travelers to plan their routes. The performance of different algorithms in terms of prediction accuracy is presented in Table 4. The results show that CNN achieves the highest prediction accuracies in all circumstances with an average prediction accuracy of 0.931, followed by OLS (0.917) and RF (0.904), which implies that it is necessary to incorporate spatiotemporal features from a network-wide perspective.

**Table 4.** Prediction performance (accuracy) of the CNN and other algorithms.

| Study Network | Model | Accuracy Score of Different Models (on Test Datasets) | | | |
|---|---|---|---|---|---|
| | | Task 1 | Task 2 | Task 3 | Task 4 |
| Network 1 | CNN | **0.939 *** | **0.942 *** | **0.925 *** | **0.928 *** |
| | OLS | 0.935 | 0.929 | 0.915 | 0.909 |
| | KNN | 0.901 | 0.897 | 0.893 | 0.890 |
| | RF | 0.917 | 0.917 | 0.910 | 0.910 |
| | ANN | 0.869 | 0.876 | 0.852 | 0.865 |
| | SAE | 0.867 | 0.870 | 0.866 | 0.866 |
| | RNN | 0.908 | 0.913 | 0.898 | 0.900 |
| | LSTM NN | 0.910 | 0.908 | 0.901 | 0.905 |
| Network 2 | CNN | **0.938 *** | **0.936 *** | **0.920 *** | **0.922 *** |
| | OLS | 0.929 | 0.920 | 0.907 | 0.897 |
| | KNN | 0.886 | 0.884 | 0.879 | 0.876 |
| | RF | 0.898 | 0.898 | 0.893 | 0.892 |
| | ANN | 0.794 | 0.867 | 0.823 | 0.832 |
| | SAE | 0.846 | 0.835 | 0.848 | 0.825 |
| | RNN | 0.901 | 0.900 | 0.896 | 0.896 |
| | LSTM NN | 0.903 | 0.907 | 0.901 | 0.895 |

Note: * indicates the best result.

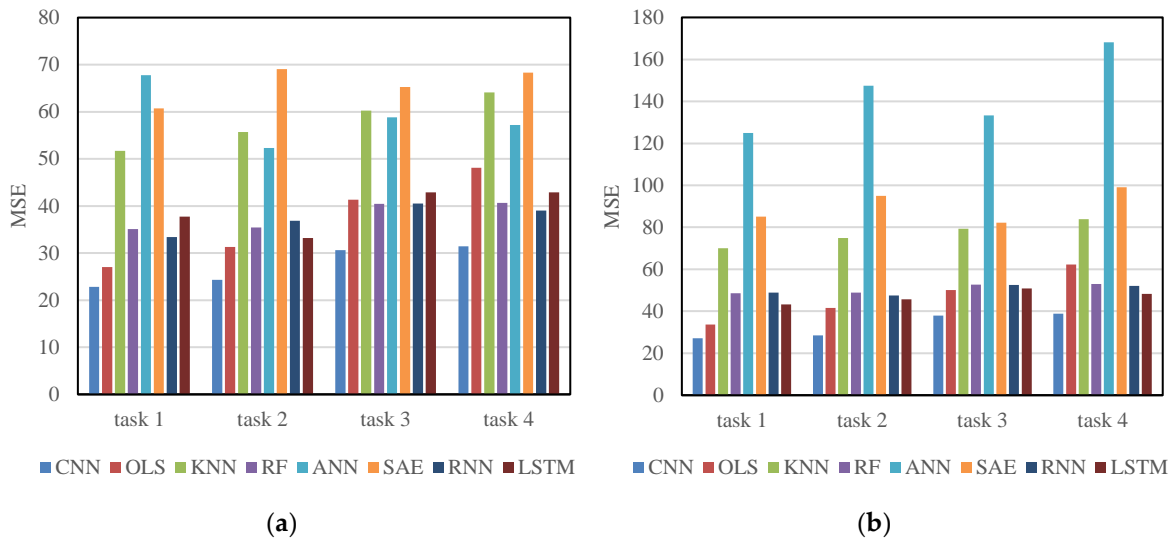

**Figure 6.** Results of different algorithms: (**a**) MSEs on Network 1; and (**b**) MSEs on Network 2.

Figure 7 shows training time of different algorithms on Networks 1 and 2. OLS, KNN, and ANN train the model more efficiently than the CNN because these algorithms have simple structures and are easy to train. However, these algorithms make significant trade-offs between their training efficiency and prediction accuracy. Other deep learning architectures, i.e., SAE, RNN and LSTM NN, require less training time than the CNN. This is primarily due to the fact that the CNN applies a large quantity of convolutional kernels to each image in order to extract extensive network-wide spatiotemporal traffic features. As for RF, it takes about nine hours to train and obtains much better results, but these results are still inferior to the CNN. RF may fail when applied to a larger-scale transportation network in real-time. Therefore, when both training efficiency and accuracy are considered, the proposed CNN outperforms the other algorithms.

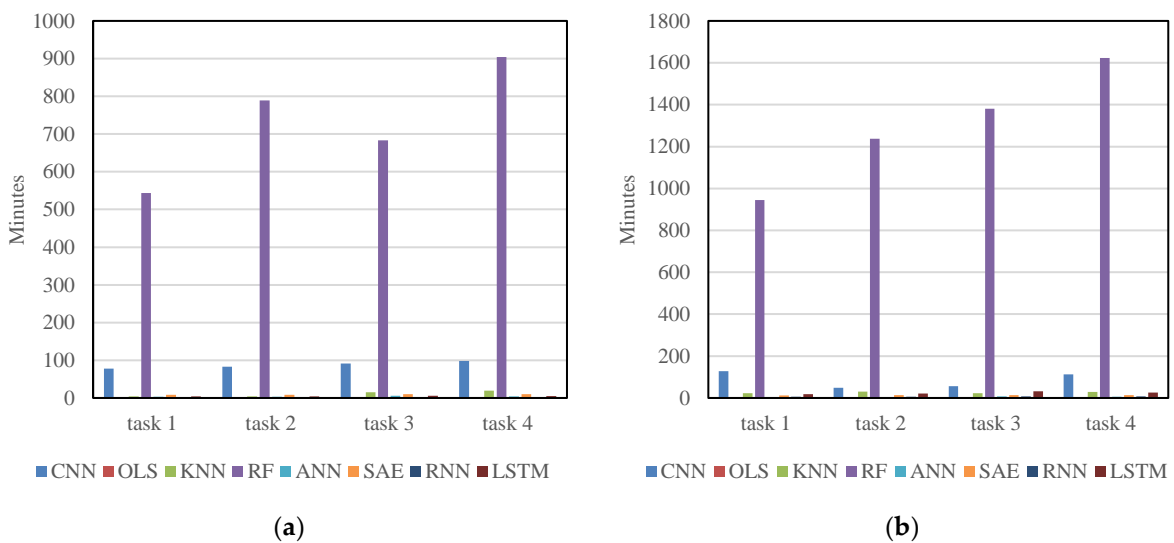

**Figure 7.** Training time of different algorithms: (**a**) training time on Network 1; and (**b**) training time on Network 2.

Based on the above discussion, useful conclusions can be yielded as follows:

- The CNN outperforms other algorithms on testing data with an average accuracy improvement of 42.91%, which implies that it is important to learn spatiotemporal features through the proposed scheme.

- The CNN trains the model within a reasonable time, but still achieves the most accurate predictions in all circumstances. As for RF, it consumes much more training time compared with CNN and receives lower accurate predictions. The OLS, KNN, and ANN train the model much faster but only yield unusable prediction results. Compared with other deep learning architectures employed, i.e., SAE, RNN and LSTM NN, the CNN trains the model much slower, but it achieves more accurate prediction results through extensive spatiotemporal features.
- The CNN performs best in long-term predictions compared with other algorithms, although making long-term traffic predictions is usually more difficult than making short-term predictions.

## 4. Conclusion

Deep learning methods are widely used in the domain of image processing with satisfactory results, since deep learning architectures usually have deeper construction and depict more complex nonlinear functions than other neural networks [25,27,30,39]. However, limited studies have addressed spatiotemporal relations among road sections in transportation networks. Spatiotemporal relations are important traffic characteristics. A better understanding of these relations will improve the accuracy of traffic prediction.

This paper proposes an image-based traffic speed prediction method that can extract abstract spatiotemporal traffic features in an automatic manner to learn spatiotemporal relations. The method contains two main procedures. The first procedure involves converting network traffic to images that represent time and space dimensions of a transportation network as two dimensions of an image. Spatiotemporal information can be preserved because surrounding road sections are adjacent in the image. The second procedure is to employ the deep learning architecture of a CNN to the image for traffic prediction. CNN has attained significant success in computer vision and performs well in image-learning tasks [31]. In this transportation prediction problem, the CNN shares the following important properties: (a) spatiotemporal features of the transportation network can be extracted automatically because of the implementation of convolutional and pooling layers of CNN; thus, the need for manual feature selection can be avoided; (b) the CNN represents network-wide traffic information of high-level features that are then used to create network-wide traffic speed predictions; and (c) the CNN can be generalized to large transportation networks because it shares weights in convolutional layers and employs the pooling mechanism. Two empirical transportation networks and four prediction tasks are considered to test the applicability of the proposed method. The results show that the proposed method outperforms OLS, KNN, ANN, RF, SAE, RNN, and LSTM NNs with an average accuracy promotion of 42.91%. The training time of the proposed method is acceptable because the proposed method achieves the best MSEs on testing data in seven (out of eight) tasks and takes much less training time than RF, which achieves the best MSEs on training data and achieves the second-best prediction accuracy on testing data.

The proposed method has some possible interesting extensions. For example, in the second procedure, other models, such as the combination of CNN and LSTM NN, would be an interesting attempt. Specifically, CNN can first extract abstract traffic features from a transportation network. The feature vectors can be fed into the LSTM NN model for prediction accuracy enhancement.

**Acknowledgments:** This work is partly supported by the National Natural Science Foundation of China (51408019, 71501009 and U1564212), Beijing Nova Program (z151100000315048), Beijing Natural Science Foundation (9172011) and Young Elite Scientist Sponsorship Program by the China Association for Science and Technology.

**Author Contributions:** Xiaolei Ma and Zhengbing He contributed analysis tools and the idea; Zhuang Dai and Jihui Ma performed the experiments and wrote the paper; Yong Wang was in charge of the final version of the paper; Yunpeng Wang collected and processed the data.

**Conflicts of Interest:** The authors declare no conflict of interest.